\documentclass[11pt]{article}

\usepackage[preprint]{acl}

\usepackage{times}
\usepackage{latexsym}

\usepackage[T1]{fontenc}

\usepackage[utf8]{inputenc}

\usepackage{microtype}

\usepackage{inconsolata}

\usepackage{graphicx}

\usepackage{algorithm}
\usepackage{algpseudocode}
\usepackage{subcaption}
\usepackage{adjustbox}
\usepackage{amsmath}
\usepackage{amssymb}
\usepackage{mathtools}
\usepackage{amsthm}
\usepackage{diagbox}
\usepackage{wrapfig}
\usepackage{float}
\usepackage{caption}
\usepackage{enumitem}
\usepackage{booktabs}
\usepackage{multirow}
\usepackage{comment}

\usepackage[ruled,noline,linesnumbered,algo2e]{algorithm2e}

\newtheorem{theorem}{Theorem}[section]

\newtheorem{definition}[theorem]{Definition}

\def\tcb{\textcolor{blue}}	

\usepackage[most]{tcolorbox}
\usepackage{xcolor} 

\newcommand\wh[1]{{\color{cyan}{#1}}}

\usepackage{listings}

\lstdefinestyle{promptstyle}{
    basicstyle=\ttfamily\small,
    breaklines=true,
    breakatwhitespace=false,
    columns=fullflexible,
    keepspaces=true,
    showstringspaces=false,
    frame=none
}

\newtcolorbox{promptbox}[1]{
    colback=gray!5,
    colframe=gray!50,
    title=\textbf{#1},
    fonttitle=\small,
    boxrule=0.4pt,
    arc=2pt,
    left=4pt,
    right=4pt,
    top=4pt,
    bottom=4pt,
    breakable
}

\usepackage[table]{xcolor}
\usepackage{tabularx}

\usepackage{pifont}

\title{Be My Tutor: On-Policy Co-Distillation \\for Mutual LLM Improvement via Peer Feedback}

\author{Woohyeon Byeon ~ Jiwon Jeon ~ Jeonghye Kim ~ Youngchul Sung\thanks{Corresponding author.} \\
  KAIST \\
  \small{\texttt{\{woohyeon.byeon, jiwon.jeon, jeonghye.kim, ycsung\}@kaist.ac.kr}} \\
  }

\begin{document}
\maketitle
\begin{abstract}
We study multi-domain LLM training in which two models, each stronger in a different domain, co-evolve by tutoring each other through on-policy feedback. Unlike one-way distillation or single-model fine-tuning, our goal is mutual Pareto improvement: each model improves across domains without losing its original strength. To this end, we propose On-Policy Co-Distillation (OPCoD), where each student's self-distillation is conditioned on its own correct rollout and feedback from its peer. To make feedback exchange effective, OPCoD uses cognizance-based gating to decide when to give feedback and feedback anchoring to ground feedback in the problem. On Science Q\&A tasks, OPCoD consistently outperforms baselines and achieves Pareto improvement across all evaluated domain pairs and students.
\end{abstract}

\section{Introduction}\label{sec:introduction}







Large language models (LLMs) are commonly fine-tuned on a single domain, such as science, medicine, or law, to acquire specialized knowledge~\citep{taylor2022galactica, singhal2025toward, hu-etal-2025-fine}. While such single-domain fine-tuning yields strong in-domain expertise, the resulting specialist often struggles beyond its domain. To broaden this coverage, recent work leverages the capacity of LLMs to absorb diverse knowledge from many domains within a single model~\citep{brown2020language, bommasani2021opportunities}, motivating multi-domain training, where datasets from different fields are combined during fine-tuning~\citep{sanh2021multitask, wei2021finetuned, chung2024scaling}.

Despite these advantages, mixing data from different domains often induces negative transfer. Gradients from one domain can interfere with those from another, degrading performance, sometimes even on the model's original specialty~\citep{cai2026advancing, yang2026disentangling, ye2026synergy}.

\begin{figure}
    \centering
    \includegraphics[width=\linewidth]{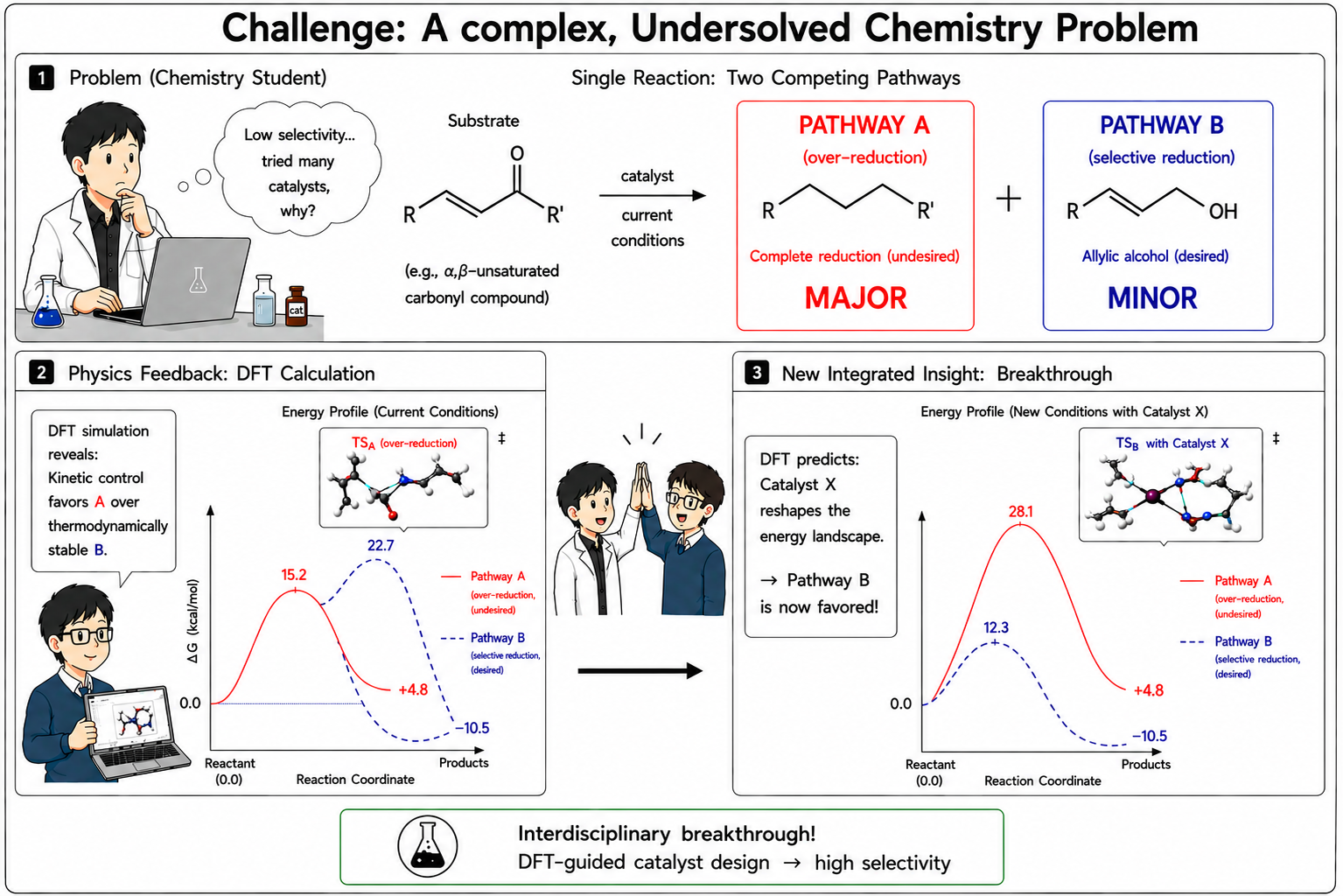}
    \caption{Conceptual illustration of OPCoD}
    \label{fig:opcd_concept}
\end{figure}

This leads us to ask: \emph{how can we leverage multi-domain training without falling into negative transfer?} To answer this, consider how humans handle a similar challenge, as illustrated in Figure~\ref{fig:opcd_concept}: a physics major and a chemistry major face an exam covering physics and chemistry, including physical chemistry at their intersection. Studying alone leaves each student's blind spots unaddressed, while tutoring each other allows them to exchange complementary knowledge and catch errors they would miss on their own. This benefit can even extend across fields: a chemistry major's chemical intuition can sometimes help a physics major solve a physics problem they could not crack alone, broadening the reasoning each can draw on.

For students to tutor each other effectively, both must exchange feedback throughout the process. Such feedback-driven learning has gained considerable attention in recent LLM research, with methods such as on-policy distillation~\citep{opd} and self-distillation~\citep{sdpo,opsd} training student models on signals from a teacher's outputs. Our setting, however, differs from these methods in two key respects: we target multi-domain capability rather than single-domain improvement, and we require bidirectional natural language feedback between two models rather than one-way teacher-to-student supervision. Motivated by these distinctions, we propose \textbf{On-Policy Co-Distillation (OPCoD)}, an on-policy co-distillation framework that enables two student models to mutually tutor each other across both domains.

\begin{figure*}[t]
    \centering
    \begin{subfigure}[b]{0.25\linewidth} 
        \centering
        \includegraphics[width=\linewidth]{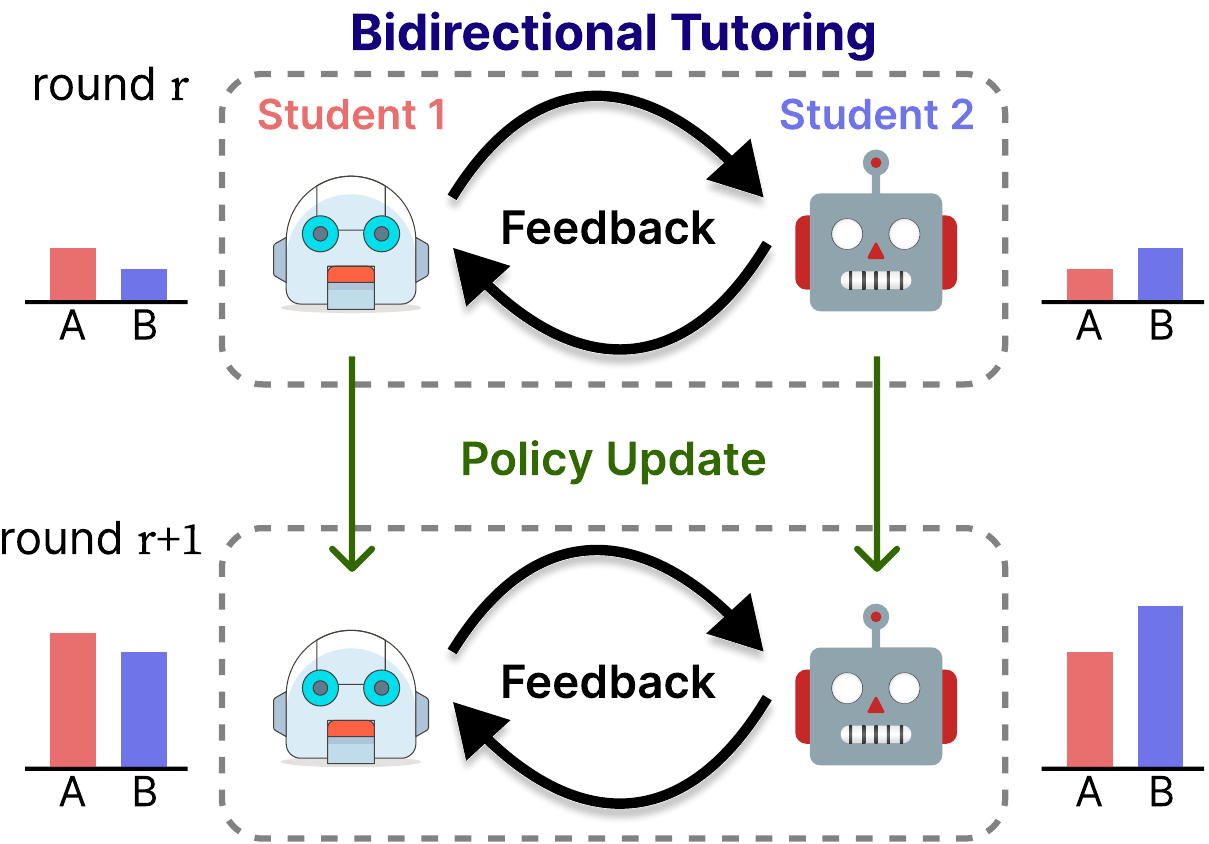}
        \label{fig:mutual_improvement}
    \end{subfigure}
    \hfill 
    \begin{subfigure}[b]{0.73\linewidth} 
        \centering
        \includegraphics[width=\linewidth]{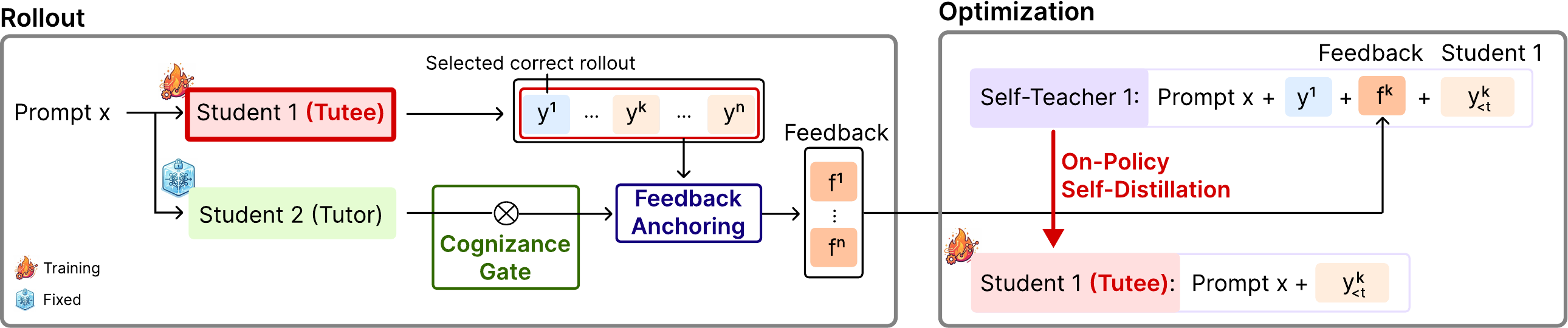} 
        \label{fig:within_round}
    \end{subfigure}
    \captionsetup{skip=0pt}
    \caption{OPCoD overview.
    \textbf{(Left)} Mutual tutoring scheme: students tutor each other through bidirectional feedback, driving policy updates.
    \textbf{(Right)} Within-round mechanism: for a prompt $x$, the tutee samples on-policy rollouts $\{y^1, \ldots, y^n\}$.
    The tutor generates feedback $\{f^1, \ldots, f^n\}$ only if it passes the cognizance gate.
    The tutee is updated by matching its policy to a self-teacher conditioned on $x$, $y^1$ (any correct rollout), and $f^k$.
    Each round applies this procedure in both directions by swapping tutee/tutor roles; for visual clarity, only Student~1 as tutee is shown.}
    \label{fig:main_combined_figure}
\end{figure*}

OPCoD realizes this through three key components: (1) \emph{co-distillation}, where the two self-distillation processes are coupled through bidirectional feedback, co-evolving throughout training; (2) \emph{cognizance-based gating}, which gates feedback by the tutor's competence across domains, mitigating negative transfer; and (3) \emph{feedback anchoring}, which grounds feedback in the specific problem to elicit informative, non-hallucinated responses. Together, these components enable mutual Pareto-improvement: both models gain capability across both domains without sacrificing performance on their original specialties.

We evaluate OPCoD on SciKnowEval~\citep{sciknoweval} across diverse multi-domain combinations (e.g., physics–chemistry, chemistry–materials), where it consistently outperforms strong baselines and achieves Pareto-improvement on both domains.
We also analyze the contribution of each design component: cognizance-based gating in preventing unreliable feedback from corrupting correct rollouts, and feedback anchoring in suppressing hallucinated feedback. Finally, we present examples of the feedback exchanged during training, illustrating how the two models tutor each other through cross-domain reasoning signals. In summary, our main contributions are:

\begin{itemize}[itemsep=0pt, topsep=0pt, parsep=0pt, partopsep=0pt, leftmargin=*]
\item We open a new direction for distillation: bidirectional between two models for multi-domain learning, rather than one-way for single-domain improvement.
\item We propose three components (\emph{co-distillation}, \emph{cognizance-based gating}, and \emph{feedback anchoring}) that enable mutual tutoring, allowing each model to improve across targeted domains without sacrificing its stronger capabilities.
\item We empirically demonstrate that OPCoD consistently achieves strong performance across diverse cross-domain combinations on SciKnowEval, outperforming baseline methods.
\end{itemize}

\section{Preliminaries}\label{sec:preliminaries}

\paragraph{On-Policy Self-Distillation.}
On-policy self-distillation (OPSD) is a training paradigm in which a single policy plays both the teacher and student roles: the student learns to match the token-level distribution of a self-teacher that has access to additional privileged information.

Let $x$ be a prompt drawn from a dataset $\mathcal{D}$ and $y=(y_1,\ldots,y_T)$ an on-policy rollout from policy $\pi_\theta$, with $y_{<t}=(y_1,\ldots,y_{t-1})$. We denote by $c$ the privileged information available only to the self-teacher. We denote the student as $\pi_S(\cdot \mid x, y_{<t}) := \pi_\theta(\cdot \mid x, y_{<t})$ and the self-teacher as $\pi_T(\cdot \mid x, c, y_{<t}) := \mathrm{sg}\!\left(\pi_\theta(\cdot \mid x, c, y_{<t})\right)$, where $\mathrm{sg}(\cdot)$ denotes the stop-gradient. 
The privileged information $c$ can take various forms, such as environment feedback, a successful rollout, or a ground-truth solution~\cite{opsd,sdpo,rlsd}.

The SDPO~\cite{sdpo} loss is
\vspace{-0.5em}
{\fontsize{9.5pt}{\baselineskip}\selectfont
\begin{equation*}\label{eq:opsd_loss}
\begin{aligned}
&\mathcal{L}_{\mathrm{SDPO}}(\pi_S)
=
\mathbb{E}_{x \sim \mathcal{D},\, y \sim \pi_S(\cdot \mid x)}
\Bigl[
\\
&\quad
{
\frac{1}{|y|}
\sum_{t=1}^{|y|}
D\!\left(
\pi_S(\cdot \mid x, y_{<t})
\,\middle\|\,
\pi_T(\cdot \mid x, c, y_{<t})
\right)}
\Bigr],
\end{aligned}
\end{equation*}
}
\vspace{-0.15em}
where $D(\cdot \,\|\, \cdot)$ is a divergence, such as KL or Jensen--Shannon.
When rich environment feedback is unavailable, SDPO uses the model's own verified correct rollout as $c$, leaving $c$ empty when no such rollout exists.

\paragraph{Pareto Criteria.}
For two evaluated domains $A$ and $B$, we say $\pi$ \emph{Pareto-dominates} $\pi'$ if $\pi$ achieves no lower score than $\pi'$ on both domains and a higher score on at least one domain~\cite{hayes2022practical}.
Given training from an initial policy $\pi_0$ to a learned policy $\pi$, we say the learned policy achieves \emph{Pareto improvement} if $\pi$ Pareto-dominates $\pi_0$.
For two students, \emph{mutual Pareto improvement} means that each learned student policy achieves Pareto improvement over its initial policy, respectively. Our goal is to achieve mutual Pareto improvement through bidirectional peer feedback, without relying on any external teacher. Formal definitions are provided in Appendix~\ref{append:pareto_definition}.

\section{OPCoD: On-Policy Co-Distillation}
\label{sec:method}

We introduce \textbf{On-Policy Co-Distillation (OPCoD)}, an on-policy co-distillation framework that drives several student models toward mutual Pareto-improvement by exchanging natural-language feedback during training. Each model performs on-policy self-distillation, with its self-teacher conditioned on both its own correct rollout and feedback from its peer. As training proceeds, each model's updates also affect its peer through the feedback it provides. The two self-distillation processes are thus coupled rather than independent, yielding a \emph{co-evolving} training dynamic.

OPCoD controls the tutor's feedback process along two complementary axes: \emph{when to give} feedback (Section~\ref{sec:gating}), via cognizance-based gating, and \emph{how to give} feedback (Section~\ref{sec:fb_anchoring}), via feedback-anchoring.
Figure~\ref{fig:main_combined_figure} illustrates the overall pipeline; Section~\ref{sec:codist} formalizes the co-distillation objective before we turn to each axis.
Pseudo-code is provided in Appendix~\ref{append:pseudo_code}.

\subsection{Co-Distillation Framework}
\label{sec:codist}

\paragraph{Problem Statement.}
We consider two domains $A$ and $B$ with their respective training sets $\mathcal{D}^A$ and $\mathcal{D}^B$. We are given two student models, $\pi^1$ and $\pi^2$.
Our goal is \emph{mutual Pareto-improvement}: both models should improve across both domains, without access to an external teacher.

\paragraph{Multi-Round Training.}
Training proceeds over $R$ rounds. Within each round, the two models alternate as \emph{tutee} and \emph{tutor}: each model is updated for $K$ on-policy self-distillation steps as a tutee, while the other model provides feedback as the tutor. Across rounds, each model's updated state shapes the feedback it provides next, realizing the coupled training dynamics.

\paragraph{Within One Round.}
Each round consists of bidirectional updates. 
First, $\pi^1$ acts as the \emph{\textbf{tutee}} and is updated for $K$ on-policy self-distillation steps using feedback from $\pi^2$ as the \emph{\textbf{tutor}}. 
Then the roles are swapped: $\pi^2$ becomes the tutee and is updated for $K$ steps using feedback from $\pi^1$. 
We describe the loss for one such directional update below; the same procedure is applied symmetrically after swapping the roles.

Let $\pi^i$ be the current tutee and $\pi^{-i}$ the current tutor ($i \in \{1, 2\}$). 
The tutee samples on-policy responses to each training prompt, and the tutor generates natural-language feedback on each response. 
The tutee then updates its student policy $\pi^i_S$ to match a self-teacher $\pi^i_T$ conditioned on two anchors that instantiate the privileged information as $c=(s,f)$, where $s$ is a correct response from the tutee's own rollouts (empty when none exists) and $f \sim \pi^{-i}(\cdot \mid x,y)$ is the tutor's feedback on the response.
For each $i\in\{1,2\}$, when $\pi^i$ serves as a tutee, it minimizes
{\fontsize{9.5pt}{\baselineskip}\selectfont
\begin{align}\label{eq:pesd_loss}
  \mathcal{L}^i&(\pi_S^i;\, \pi^{-i})  
  = \mathbb{E}_{x \sim \mathcal{D},\; y \sim \pi_S^i(\cdot \mid x)}
     \!\Big[ \\[-0.6em]
     & \frac{1}{|y|}\sum_{t=1}^{|y|} D\Big(\,
     \pi_S^i(\cdot \mid x, y_{<t})
     \,\Big\|\,
     \pi_T^i(\cdot \mid x,\, s, \,f, \,y_{<t})
     \Big)\Big], \notag
\end{align}
}
where $D$ is a divergence (e.g., Jensen--Shannon).



\begin{figure*}[t]
    \centering
    \includegraphics[width=\linewidth]{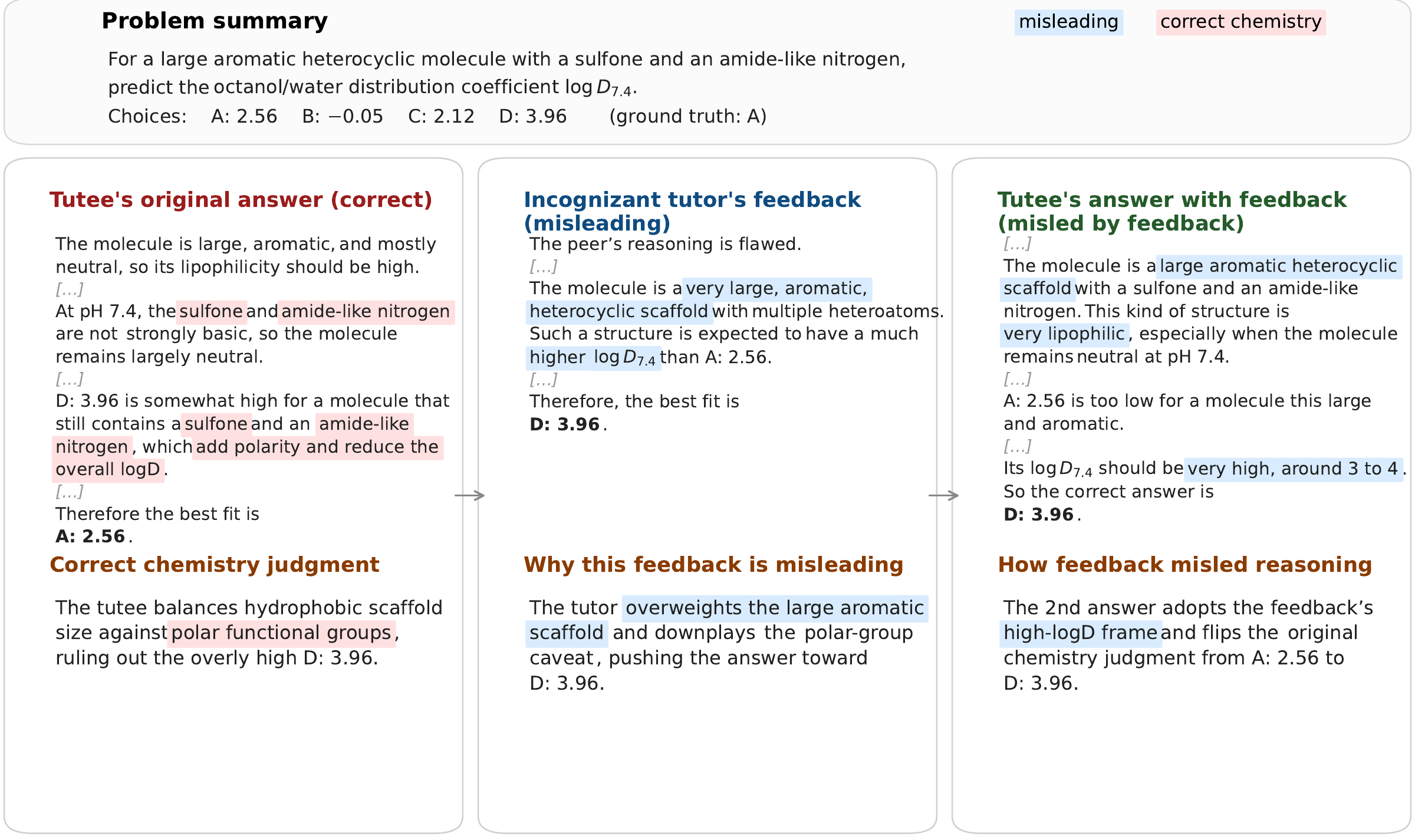}
    \caption{\textbf{Necessity of cognizance-based gating: }Incognizant tutor's feedback can break an initially correct answer.
The tutee first rules out the overly high $\log D_{7.4}$ value by considering polar functional groups, but the incognizant tutor's feedback overemphasizes the large aromatic scaffold.
The second answer then adopts this high-$\log D$ frame and flips from the correct choice A to the incorrect choice D.
}
    \label{fig:incog_break_example}
\end{figure*}

\begin{figure*}[t]
    \centering
    \includegraphics[width=\linewidth]{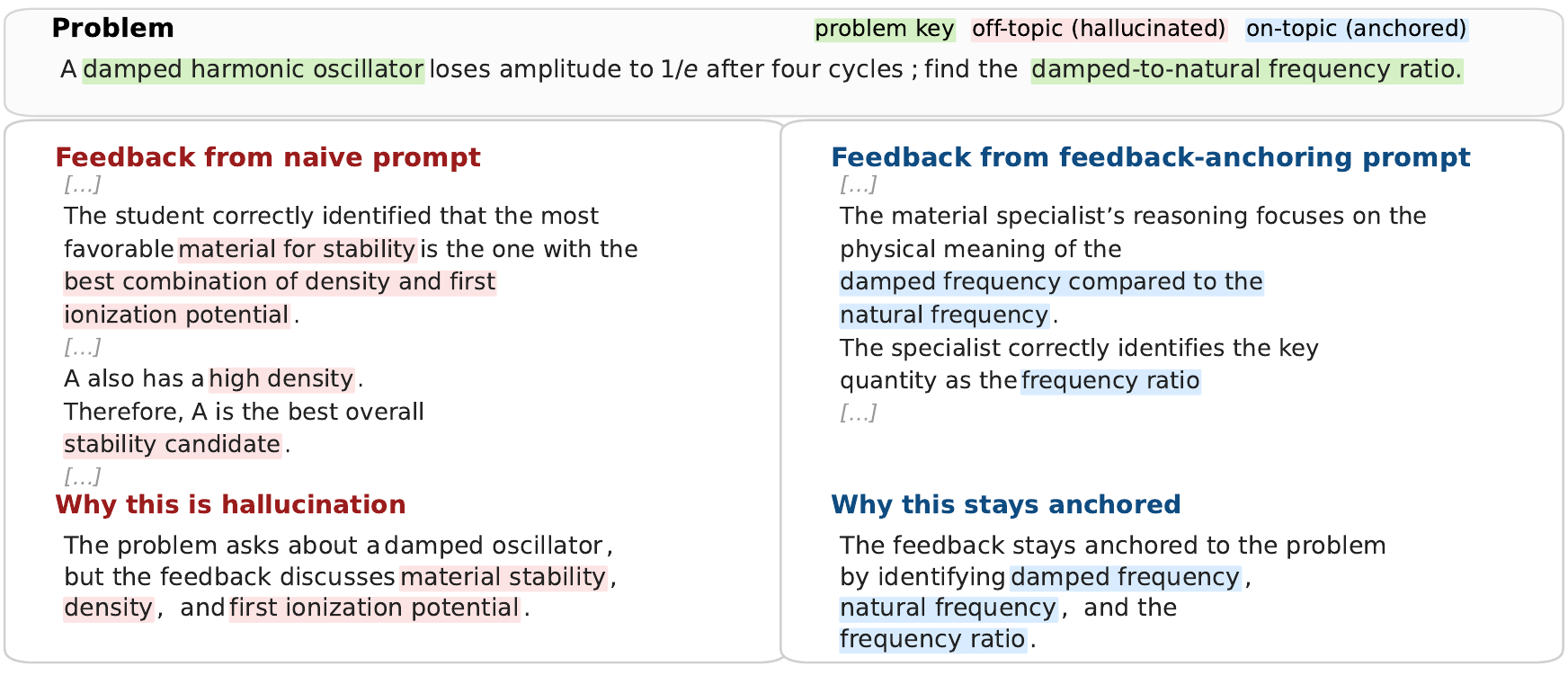}
    \caption{\textbf{Necessity of feedback anchoring:} Example of feedback hallucination and feedback anchoring. The left panel shows problem-irrelevant feedback generated by the naive prompt, while the right panel shows feedback generated by our anchoring prompt, which stays grounded in the problem.}
    \label{fig:anchor_demo}
\end{figure*}

\subsection{When to Give Feedback: Cognizance-Based Gating}
\label{sec:gating}

Not all tutor feedback helps the tutee. A tutor that is insufficiently reliable in the problem's relevant area may mislead the self-teacher and degrade the distillation signal. Figure~\ref{fig:incog_break_example} illustrates such a case, where inappropriate tutor feedback corrupts the response of the tutee's self-teacher and can mislead the tutee's learning. We therefore let the tutor give feedback only when it is sufficiently \emph{cognizant of the relevant domains}, as measured by its relative performance gap.

\paragraph{Cognizance-Based Gating.}
At the start of each round, both $\pi^1$ and $\pi^2$ are evaluated on held-out validation sets for each domain. Let $s^i_d$ denote model $\pi^i$'s validation score on domain $d$, and let $s^*_d = \max\{s^1_d, s^2_d\}$ denote the highest score on domain $d$. To evaluate the suitability of each $\pi^i$ as a tutor, we define the \emph{cognizance gap} of $\pi^i$ as the total relative shortfall from $s^*_d$: for $i=1,2$,
\begin{equation}\label{eq:cognizance_gap}
\Delta^i = \sum_{d \in \{A, B\}} \frac{s^*_d - s^i_d}{s^*_d}.
\end{equation}
Intuitively, each term measures the $\pi^i$'s relative gap from the best model on a domain, so a smaller $\Delta^i$ means $\pi^i$ is broadly closer to the best across domains.
We consider $\pi^i$ \emph{cognizant} if $\Delta^i \leq \tau$ and \emph{incognizant} otherwise, where $\tau$ is a predefined threshold. 
At the start of each round, if $\pi^i$ is cognizant, it gives feedback as a tutor for this round, providing the privileged information $c=(s,f)$ for the tutee's OPCoD update (Eq.~\eqref{eq:pesd_loss}); if $\pi^i$ is incognizant, it gives no feedback for this round, and the tutee falls back to self-distillation ($c=s$). This criterion is relative rather than absolute: it asks whether the tutor is likely to add value beyond what the tutee can already obtain on its own.

For example, if two models have scores $(s^1_A,s^1_B)=(100,70)$ and $(s^2_A,s^2_B)=(70,100)$ with $\tau=0.2$, both have $\Delta=0.3$ and give no feedback as tutors for each other: although a score of $70$ indicates a reasonably capable tutor in absolute terms, the tutee already has stronger self-generated rollouts on its own high-scoring domain.

\subsection{How to Give Feedback: Feedback-Anchoring}
\label{sec:fb_anchoring}

A natural starting point for feedback generation is to prompt the tutor with a generic instruction: read the question and the tutee's response, then provide constructive feedback. 
In practice, this can produce feedback disconnected from the actual problem, addressing irrelevant content rather than the tutee's specific reasoning. 
We refer to this phenomenon as \emph{feedback hallucination}. 
Figure~\ref{fig:anchor_demo} illustrates this failure mode and how anchoring mitigates it.

\paragraph{Feedback Anchoring.}
Feedback anchoring uses a two-step prompt. 
First, the tutor must identify a single technical concept from the question and output it in \texttt{<concept>}...\texttt{</concept>} tags. 
Second, the tutor writes a short critique of the tutee's reasoning without revealing the final answer. 
The extracted concept also serves as a verification signal: if it does not appear explicitly in the question text, we discard the feedback as ungrounded. 
Verified feedback is then sanitized to remove direct answer reveals, and the concept tag is stripped before the feedback is given to the tutee. 
The full prompt and filtering details are provided in Appendix~\ref{append:feedback_anchoring}.

\section{Experiments}\label{sec:experiments}


We organize our experiments around four aspects of OPCoD. 
First, we evaluate whether co-distillation achieves Pareto improvement across paired domains. 
Second, we examine whether cognizance-based gating mitigates the risk that feedback corrupts previously correct rollouts.
Third, we analyze whether feedback anchoring preserves enough tutor feedback while filtering problem-irrelevant responses. 
Finally, we examine how peer feedback can supply complementary reasoning insights that help a student correct its own mistake.

\subsection{Experimental Setting}\label{sec:exp_setting}

We use the Science Q\&A subset from the L3 split of SciKnowEval~\cite{sciknoweval}, which spans four scientific domains: chemistry, physics, materials science, and biology. Following~\citet{sdpo}, we partition each domain's data into training and test splits.

In our experiments, we use three students based on Qwen3-8B~\cite{yang2025qwen3} with different domain strengths, whose construction details are deferred to Appendix~\ref{append:experimental_detail}.
To cover a range of co-training scenarios, we pair two students from distinct domains and jointly train them on the union of their respective training splits. Because the biology split contains substantially fewer examples than the other three, we focus on three pairs that exclude biology: chemistry--materials, materials--physics, and physics--chemistry. 

As baselines, we train each student individually on the same union dataset via GRPO~\cite{grpo} and SDPO~\cite{sdpo}, isolating the effect of peer feedback by contrasting solo and joint training under matched data. 

Each student is trained for 100 update steps. 
We use $n=8$ rollouts per prompt and the cognizance threshold $\tau=0.2$. 
We report avg@16 on the final checkpoint; remaining hyperparameters, including the divergence and learning rate, are provided in Appendix~\ref{append:hyperparams}.





\subsection{Results: Multi-Domain Science Q\&A}\label{sec:results}

\begin{table*}[t]
\centering
\renewcommand{\arraystretch}{0.9}
\setlength{\tabcolsep}{4pt}
\begin{tabularx}{\textwidth}{l*{9}{>{\centering\arraybackslash}X}}
\toprule
 & \multicolumn{3}{c}{\textbf{Mat-Phys}} & \multicolumn{3}{c}{\textbf{Chem-Mat}} & \multicolumn{3}{c}{\textbf{Phys-Chem}} \\
\cmidrule(lr){2-4} \cmidrule(lr){5-7} \cmidrule(lr){8-10}
 & Mat. & Phys. & Avg. & Chem. & Mat. & Avg. & Phys. & Chem. & Avg. \\
\midrule
 & \multicolumn{3}{c}{\textit{Mat-stronger}} & \multicolumn{3}{c}{\textit{Chem-stronger}} & \multicolumn{3}{c}{\textit{Phys-stronger}} \\
Student 1 & 65.1 & 48.1 & 56.6 & 56.3 & 56.1 & 56.2 & 51.6 & 37.3 & 44.5 \\
+ GRPO   & 67.8 & 53.8 & 60.8 & 62.0 & 65.6 & 63.8 & 52.6 & 55.4 & 54.0 \\
+ SDPO   & 62.2 & 52.3 & 57.2 & 70.9 & 65.8 & 68.4 & 51.1 & 56.7 & 53.9 \\
\rowcolor{blue!10}
+ OPCoD   & \textbf{70.5} & \textbf{55.3} & \textbf{62.9} & \textbf{71.3} & \textbf{69.8} & \textbf{70.6} & \textbf{54.1} & \textbf{58.9} & \textbf{56.5} \\
\midrule
 & \multicolumn{3}{c}{\textit{Phys-stronger}} & \multicolumn{3}{c}{\textit{Mat-stronger}} & \multicolumn{3}{c}{\textit{Chem-stronger}} \\
Student 2 & 56.0 & 51.6 & 53.8 & 37.6 & 65.1 & 51.4 & 48.6 & 56.3 & 52.5 \\
+ GRPO   & 66.0 & 53.9 & 60.0 & 57.3 & 65.2 & 61.2 & 54.3 & 62.8 & 58.6 \\
+ SDPO   & 58.8 & 51.9 & 55.4 & 55.4 & 60.0 & 57.7 & 56.4 & 69.9 & 63.2 \\
\rowcolor{blue!10}
+ OPCoD   & \textbf{66.1} & \textbf{54.3} & \textbf{60.2} & \textbf{57.5} & \textbf{66.4} & \textbf{62.0} & \textbf{58.8} & \textbf{70.2} & \textbf{64.5} \\
\bottomrule
\end{tabularx}
\caption{Avg@16 results across three domain pairs. OPCoD is highlighted. Each block label (e.g., Mat-stronger) indicates the student with the highest score in that domain among the three initial students.}
\label{tab:codistill_combined_qwen3_8b}
\end{table*}

Table~\ref{tab:codistill_combined_qwen3_8b} reports avg@16 scores across the three domain pairs. OPCoD achieves \emph{mutual Pareto improvement} in every pair: both models in each pair improve on both domains.
Moreover, OPCoD \emph{Pareto-dominates} the GRPO and SDPO baselines, achieving the highest score in every column.

Mutual Pareto improvement, our main goal, is not achieved by all baselines. 
In SDPO runs, the per-pair average increases because gains on the non-native domain offset losses on the native domain. 
However, SDPO degrades native-domain performance in three of the six (agent, pair) configurations: the Mat-stronger agent's Mat score drops from 65.1 to 62.2 in Mat--Phys and from 65.1 to 60.0 in Chem--Mat, and the Phys-stronger agent's Phys score drops from 51.6 to 51.1 in Phys--Chem. 
This is exactly the negative-transfer pattern: independent training on the union of domains can improve a model's non-native domain at the cost of its original specialty. 
OPCoD's gating prevents a tutor from giving feedback when it is not sufficiently reliable, eliminating this failure mode while still delivering strong non-native-domain gains.

\newcommand{\gain}[1]{\textsubscript{\textcolor{red!70!black}{(#1)}}}

\subsection{Cognizance-Based Gating Mitigates the Risk of Feedback Corruption}
\label{sec:break-risk}

Tutor feedback directly affects training through the self-teacher's conditioning (Eq.~\eqref{eq:pesd_loss}); misleading feedback can therefore corrupt the student's training signal. 
To validate that cognizance-based gating mitigates this risk, we define the \emph{break-rate} for each (tutee, tutor) pair as the fraction of the tutee's previously correct rollouts that become incorrect after the tutor's feedback is added to the self-teacher's conditioning. 
Across the three domain pairs, incognizant tutors (which gating excludes) break correct rollouts at $2.4\times$ the rate of cognizant tutors (which gating admits), confirming the effectiveness of gating's selection rule.
Detailed settings and aggregate break-rates are provided in Appendix~\ref{append:break-risk-appendix}.


Gating's benefit is robust across problem difficulty. 
As shown in Table~\ref{tab:break-risk-difficulty}, cognizant tutors have lower break-rates than incognizant tutors in every difficulty range. 
The gap appears even on easier problems ($1.13\%$ vs. $4.75\%$ on the easiest range) and remains substantial on harder ones ($5.42\%$ vs. $13.73\%$ on the hard range and $13.79\%$ vs. $21.88\%$ on the hardest). 
Gating's safety advantage is therefore consistent across difficulty levels, and especially valuable on harder problems where correct rollouts are scarce.


A natural alternative to our gating rule (which silences all incognizant feedback for the round) is to admit feedback from an incognizant tutor only for problems in its stronger domain. 
However, even within the tutor's stronger domain, incognizant tutors still break correct rollouts at $1.4\times$ the rate of cognizant tutors. 
This indicates that domain restriction alone is insufficient: incognizant feedback remains substantially riskier than cognizant feedback. 
We evaluate this domain-selective alternative as an ablation in Figure~\ref{fig:gating_ablation_nonum}, further supporting our gating rule in Section~\ref{sec:gating}.

\newcommand{\ratio}[1]{ \textsubscript{\textcolor{red!70!black}{(#1)}}} 

\begin{table}[H]
\centering
\small
\setlength{\tabcolsep}{7pt}
\renewcommand{\arraystretch}{1.15}
\begin{tabular}{lcc}
\toprule
\textbf{Difficulty} & \textbf{Cognizant (\%)} & \textbf{Incognizant (\%)} \\
\midrule
Easiest  & 1.13  & 4.75\ratio{$\times$4.22}  \\
Easy        & 2.35  & 5.54\ratio{$\times$2.36}  \\
Hard        & 5.42  & 13.73\ratio{$\times$2.53} \\
Hardest       & 13.79 & 21.88\ratio{$\times$1.59} \\
\bottomrule
\end{tabular}
\caption{Break-rate by difficulty, defined by incorrect pre-feedback rollouts among 8: very easy (0--1), easy (0--4), hard (5--7), and hardest (7).}
\label{tab:break-risk-difficulty}
\end{table}

\subsection{Feedback Anchoring Suppresses Hallucinations}\label{sec:analysis_fb_anchoring}

\begin{figure*}[t]
    \centering
    \includegraphics[width=\linewidth]{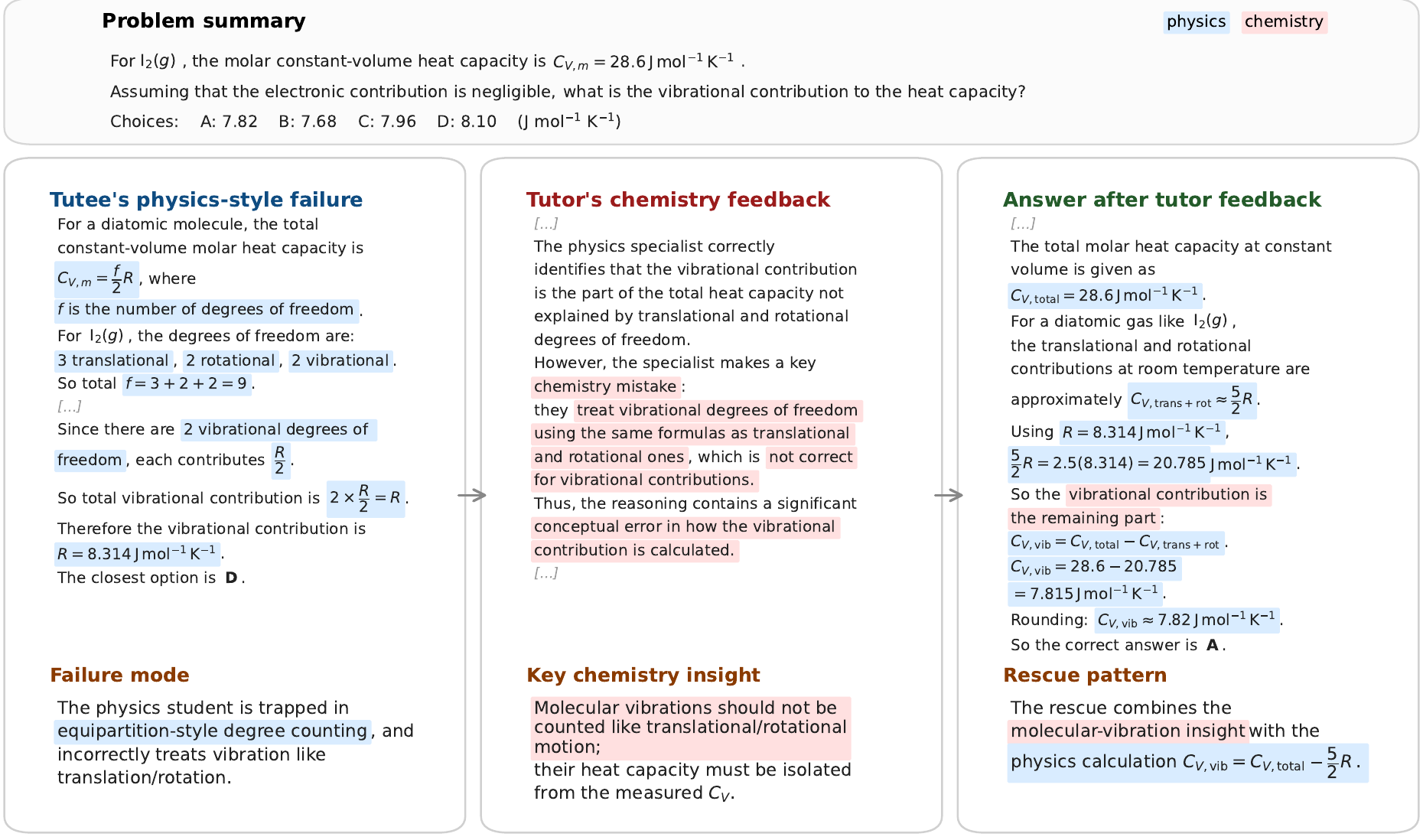}
    \caption{Case study of peer tutoring. \textbf{Left}: a physics-stronger tutee attempts a heat-capacity problem and reaches an incorrect answer through physics-style equipartition reasoning. \textbf{Middle}: a chemistry-stronger, physics-cognizant tutor localizes the error and supplies the missing physical-chemistry insight without revealing the answer. \textbf{Right}: given the same problem with the feedback, the tutee revises its reasoning by combining its physics fluency with the tutor's physical-chemistry insight to reach the correct answer.}
    \label{fig:case_study_phys_150}
\end{figure*}



\begin{figure}[H]
    \centering
    \includegraphics[width=\linewidth]{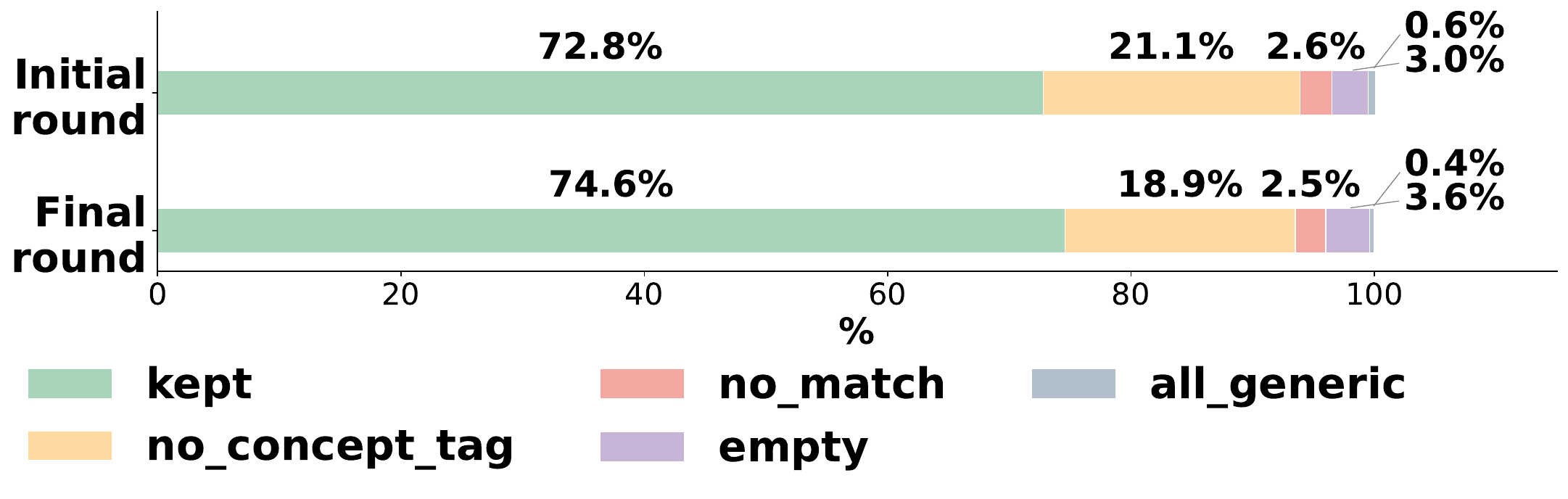}
    \captionsetup{skip=0pt}
    \caption{Filtering statistics under feedback anchoring at the initial and final training rounds.}
    \label{fig:feedback_filtering_stacked_bar}
\end{figure}

Figure~\ref{fig:feedback_filtering_stacked_bar} reports the breakdown of filter outcomes for tutor-generated feedback under feedback anchoring at the initial and final rounds of training. As described in Section~\ref{sec:fb_anchoring}, the anchoring prompt requires the tutor to extract a key concept from the question in \texttt{<concept>} tags. We then classify each feedback by \emph{rule-based string matching} into one of five categories. \emph{Kept} means the extracted concept appears explicitly in the question text, while \emph{no\_match} means it does not, signaling ungrounded feedback. The remaining categories (\emph{no\_concept\_tag}, \emph{empty}, and \emph{all\_generic}) are cases where the concept tag is absent, empty, or too generic to verify against the question, so we discard them to prioritize precision over recall. Detailed criteria for each category are in Appendix~\ref{append:sanitize_fb}.

Two observations stand out. First, the kept rate stays consistently above 70\% and slightly increases from 72.8\% to 74.6\%, showing that the anchoring filter preserves sufficient training signal. Second, \emph{no\_match} remains a rare category at about 2.5\%, suggesting that feedback anchoring effectively suppresses problem-irrelevant feedback.


\vspace{-2em}


\subsection{A Case Study of Cross-Domain Tutoring}\label{sec:case_study}

Figure~\ref{fig:case_study_phys_150} illustrates how co-training can help a student recover even when the error lies within its stronger domain.
The problem is a physics-labeled heat-capacity question, and the tutee is the student that is stronger in physics than the tutor. 
Nevertheless, the tutee initially answers incorrectly. 
The tutor is stronger in chemistry, and its feedback supplies a complementary physical-chemistry perspective that fills a gap in the tutee's reasoning.

The tutee's initial reasoning is physics-style in the sense that it relies on equipartition and degree-of-freedom counting. 
It counts translational, rotational, and vibrational degrees of freedom, then treats the vibrational part as if each vibrational degree contributed in the same way as translation or rotation. 
Thus, the failure is not that the tutee ignores vibration, but that it handles molecular vibration with the wrong heat-capacity interpretation.

The tutor feedback identifies this issue without revealing the answer. Its key insight is that the vibrational contribution should be isolated from the measured heat capacity, rather than counted like translational or rotational motion. With this feedback, the tutee combines the tutor's molecular-vibration insight with its own physics fluency: it treats the vibrational contribution as the remainder of the measured heat capacity after accounting for translational and rotational motion, thereby recovering the correct answer.

This example illustrates why co-training diverse students can be useful. 
Even when a problem lies in one student's stronger domain, another student may supply a complementary perspective that pinpoints the reasoning gap and enables the original student to complete the reasoning correctly.
Additional examples are provided in Appendix~\ref{append:additional_examples}.

\section{Ablation }\label{sec:ablation}



\begin{figure}[H]
    \centering
    \includegraphics[width=0.9\linewidth]{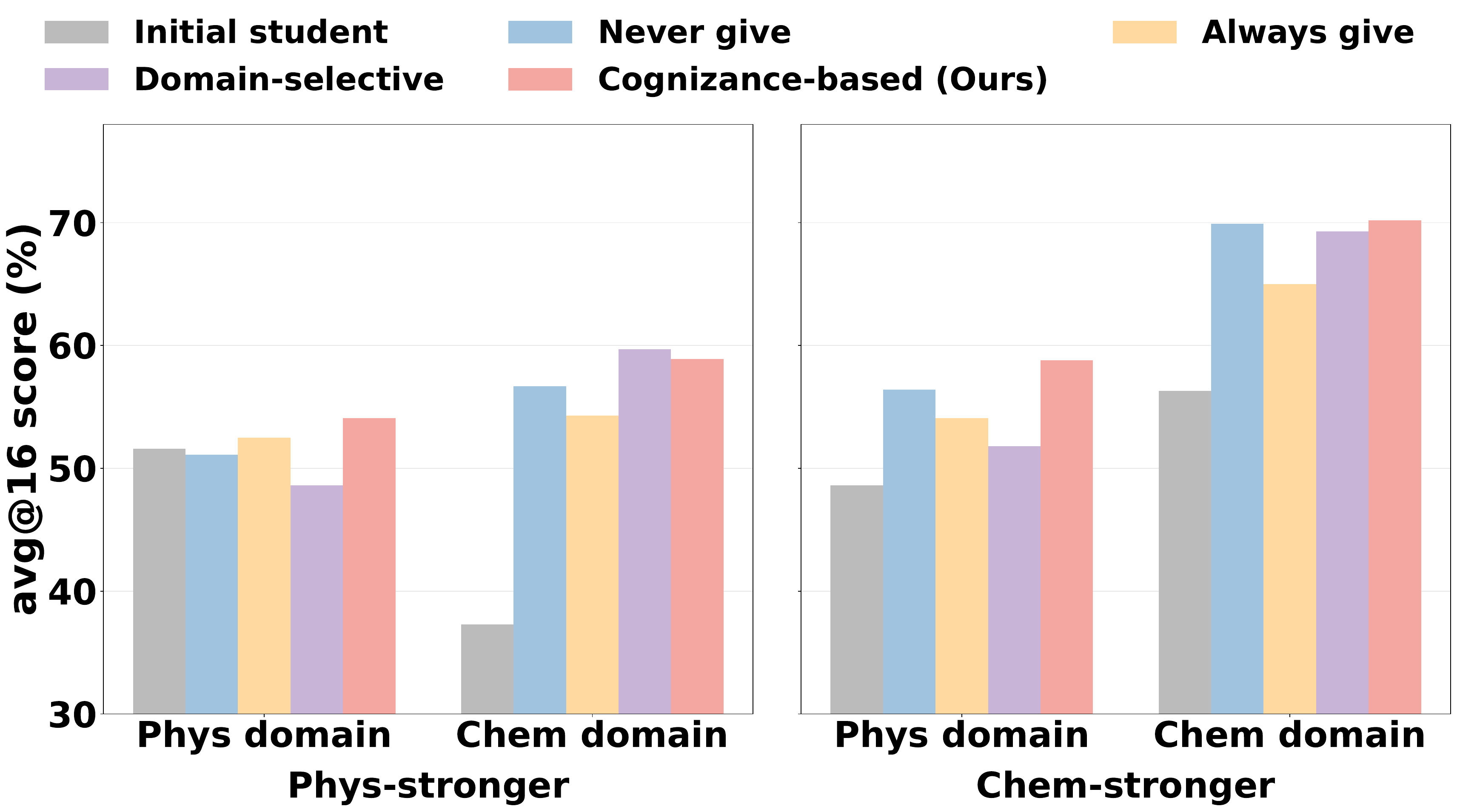}
    \captionsetup{skip=0pt}
    \caption{Feedback gating strategy ablation on the physics--chemistry pair, shown for the physics-stronger student (left) and chemistry-stronger student (right).}
    \label{fig:gating_ablation_nonum}
\end{figure}

\paragraph{Cognizance-Based Gating Ablation.}
We ablate when tutor feedback is given. Figure~\ref{fig:gating_ablation_nonum} compares four gating strategies on the physics--chemistry pair: \emph{Always give}, \emph{Never give}, \emph{Domain-selective}, and \emph{Cognizance-based} gating. \emph{Domain-selective} gating still allows feedback from an incognizant tutor only on its stronger domain, testing whether a tutor's expertise can compensate for its overall incognizance (Section~\ref{sec:break-risk}). \emph{Never give} and \emph{Domain-selective} improve the chemistry score of the physics-stronger student, but reduce its physics score below the initial score, showing negative transfer and failing Pareto improvement. Although \emph{Always give} achieves Pareto improvement, \emph{Cognizance-based} gating Pareto-dominates it and achieves Pareto improvement with higher scores.

\begin{figure}[H]
    \centering
    \includegraphics[width=0.9\linewidth]{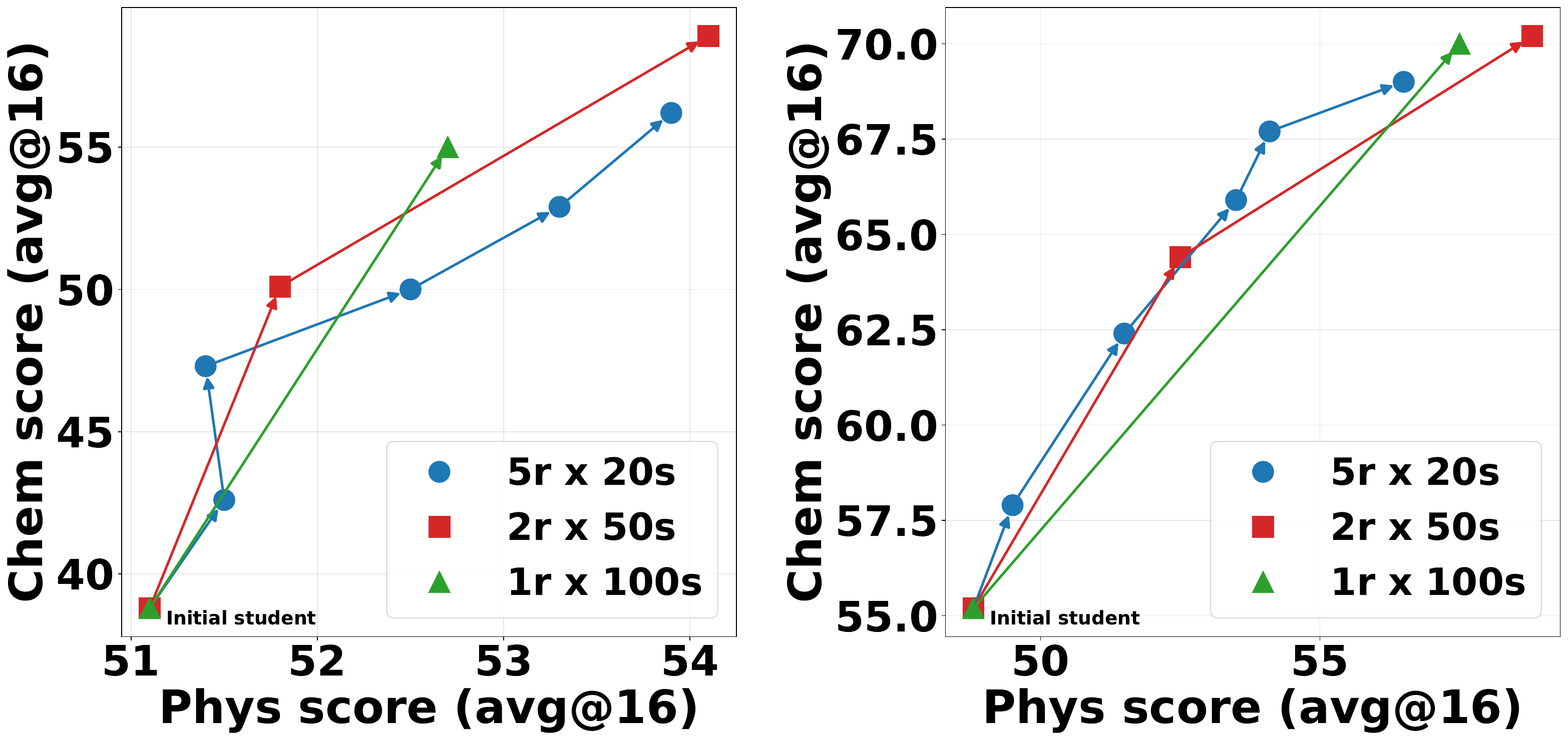}
    \captionsetup{skip=0pt}
    \caption{Round-step ablation on the physics--chemistry pair, showing evaluation trajectories: physics-stronger student (left), chemistry-stronger student (right).}
    \label{fig:round_step_path}
\end{figure}

\paragraph{Round-Step Ablation.}
We ablate how to allocate a fixed number of OPCoD training steps across rounds. 
Figure~\ref{fig:round_step_path} compares $5\times20$, $2\times50$, and $1\times100$ schedules on the physics--chemistry pair, where $r\times s$ denotes $r$ rounds with $s$ training steps per round. 
All schedules improve over the initial student, showing robustness to round-step choice.
The $2\times50$ schedule yields strong trajectories for both students, suggesting that both multi-round training (which $1\times100$ lacks) and sufficient within-round updates (which $5\times20$ lacks) are important.

\section{Related Works}\label{sec:RelatedWorks}

\textbf{Multi-Domain LLMs. }
Multi-domain training improves cross-task performance~\citep{wei2021finetuned, chung2024scaling}, but can induce negative transfer that weakens domain-specific expertise. 
Existing approaches mitigate this by modifying gradients, adapters, or training schedules~\citep{cai2026advancing, yang2026disentangling, ye2026synergy}, while multi-teacher distillation aggregates supervision from several teachers into a single student~\citep{xiao2026mimo,yang2026nemotron}. 
In contrast, OPCoD addresses negative transfer through co-distillation with cognizance-based gating, rather than one-way transfer into a single model, without relying on external teachers.

\noindent\textbf{On-Policy Self-Distillation. }
On-policy self-distillation trains an LLM from its own rollouts, with the same model acting as both student and self-teacher. 
Recent methods condition the self-teacher on privileged information such as successful rollouts, environment feedback, or richer feedback signals~\citep{opsd,sdpo,kim2026rebellious,song2026expanding,ye2026policy}. 
Unlike these single-model methods, OPCoD couples two self-distillation processes through bidirectional feedback in a multi-student, multi-domain setting.

\noindent\textbf{Multi-LLM Co-Training. }
Existing work trains multiple LLM agents mainly to improve inference-time collaboration, such as debate, verifier-scored discussion, or persuasion-balanced dialogue~\citep{maporl,liao2025marft,stengel2025teaching,subramaniam2025multiagent}. 
In contrast, OPCoD uses peer interaction only during training: students exchange feedback to transfer complementary reasoning signals, but are evaluated independently across all paired domains.


\section{Conclusion}\label{sec:conclusion}

We presented OPCoD, an on-policy co-distillation framework where student LLMs improve together through peer feedback.
In OPCoD, each student performs on-policy self-distillation with a self-teacher conditioned on both its own correct rollout and peer feedback, where the feedback is controlled by cognizance-based gating and feedback anchoring.
On Science Q\&A tasks, OPCoD achieves mutual Pareto improvement across all evaluated domain pairs, outperforming baselines.
Our analyses show that properly gated and anchored feedback can provide complementary reasoning cues while avoiding unreliable or problem-irrelevant feedback.



\section*{Limitations}


First, our experiments focus on Science Q\&A tasks, and it remains unknown whether the method is effective for more distant, non-science domain combinations.
Second, the current approach is restricted to pairwise co-distillation between two students, and settings involving more than two agents remain unexplored.
Third, the framework relies on prompt-based feedback generation, meaning that the detailed wording of the prompt may still affect feedback quality and downstream performance.

\section*{Potential Risks}
Because OPCoD trains students using feedback generated by other students, its reliability depends on the trustworthiness of the participating models. 
If a malicious or compromised student is included, it may introduce misleading feedback that is later distilled into another student. 
This motivates using trusted participants, feedback validation, and monitoring when applying OPCoD beyond controlled experimental settings.

\bibliography{custom}

\newpage
\appendix

\clearpage

\section{Formal Definitions of Pareto Criteria}
\label{append:pareto_definition}

Consider two students indexed by $i \in \{1, 2\}$ and two domains $d \in \{A, B\}$. Let $s_d(\pi)$ denote the evaluation score of policy $\pi$ on domain $d$, where higher scores indicate better performance.

\begin{definition}[Pareto dominance]
A policy $\pi$ \emph{Pareto-dominates} a policy $\pi'$, denoted $\pi \succ_{\mathrm{P}} \pi'$, if
\[
\begin{aligned}
s_d(\pi) &\ge s_d(\pi'), \quad \forall d \in \{A, B\}, \text{ and}\\
s_d(\pi) &> s_d(\pi'), \quad \exists d \in \{A, B\}.
\end{aligned}
\]
\end{definition}

\begin{definition}[Pareto improvement]
Training from an initial policy $\pi_0$ to a learned policy $\pi$ \emph{achieves Pareto improvement} if
\[
\pi \succ_{\mathrm{P}} \pi_0.
\]
\end{definition}

\begin{definition}[Mutual Pareto improvement]
Training from initial policies $\pi_0^1, \pi_0^2$ to learned policies $\pi^1, \pi^2$ \emph{achieves mutual Pareto improvement} if
\[
\pi^i \succ_{\mathrm{P}} \pi_0^i, \quad \forall i \in \{1, 2\}.
\]
\end{definition}

\newpage
\section{Pseudo-Code}\label{append:pseudo_code}

\begin{algorithm}[H]
\DontPrintSemicolon
\SetKwIF{If}{ElseIf}{Else}{if}{}{else if}{else}{end if}
\caption{On-Policy Co-Distillation (OPCoD)}
\label{alg:opcod}

\BlankLine
\textbf{Input:} Initial students $\pi_1^1,\pi_1^2$, threshold $\tau$, rounds $R$, steps $K$\;
\BlankLine

\For{$r=1,\dots,R$}{
    Compute cognizance gaps $\Delta_r^1,\Delta_r^2$ at round $r$ using
    validation scores~(Eq.~\eqref{eq:cognizance_gap})\\
    $\bar{\pi}_r^1 \leftarrow \pi_r^1,\quad \bar{\pi}_r^2 \leftarrow \pi_r^2$ \tcp*{Frozen tutors}

    \BlankLine
    \For{$i\in\{1,2\}$}{
        $\pi_{\mathrm{tutee}} \leftarrow \pi_r^i$,\;
        $\pi_{\mathrm{tutor}} \leftarrow \bar{\pi}_r^{-i}$\;

        \BlankLine
        \For{$k=1,\dots,K$}{
            \ForEach{$x\in\mathcal{D}$}{
                $y^1,\dots,y^n \sim \pi_{\mathrm{tutee}}(\cdot\mid x)$\\
                $s \leftarrow$ correct rollout in $\{y^j\}_{j=1}^n$, if any; otherwise $\emptyset$\;

                \uIf(\tcp*[f]{Cognizant}){$\Delta_r^{tutor}\le\tau$}{
                    Generate and sanitize feedback using
                    feedback anchoring (Appendix~\ref{append:feedback_anchoring})\\
                    $f^j \sim \pi_{\text{tutor}}(\cdot|x,y^j),~\forall j$\\
                    $f^j \leftarrow \text{Sanitize}(f^j),~\forall j$\;
                }
                \Else(\tcp*[f]{Incognizant}){
                    $f^1,\dots,f^n \leftarrow \emptyset$\;
                }
            }

            Update $\pi_{\mathrm{tutee}}$ using $\mathcal{L}^i(\pi_{\mathrm{tutee}};\pi_{tutor})$ (Eq.~\eqref{eq:pesd_loss})\;
        }

        $\pi_{r+1}^i \leftarrow \pi_{\mathrm{tutee}}$\;
    }
}
\BlankLine
\textbf{Return:} $\pi_{R+1}^1,\pi_{R+1}^2$\;
\end{algorithm}

\section{Feedback Anchoring Process}\label{append:feedback_anchoring}

\subsection{Prompts for Feedback Generation}\label{append:prompt}

\begin{promptbox}{Prompt for feedback generation with generic instruction}\label{append:default_prompt}
\begin{lstlisting}[style=promptstyle]
You are a domain expert reviewing a student's response to a conversation or problem.
Read the full context below, then evaluate the student's final response. \
Provide concise, constructive feedback: point out what is correct, what is wrong \
or missing, and how the student should improve their reasoning.
Do NOT directly reveal the correct answer.

Context:
{problem}

Student's response:
{response}

Expert feedback:
\end{lstlisting}
\end{promptbox}

\begin{promptbox}{Prompt for feedback generation with feedback anchoring}\label{append:fb_anchoring_prompt}
\begin{lstlisting}[style=promptstyle]
You are a {peer_domain} reasoning specialist. You excel at {peer_domain} concepts and {peer_domain} reasoning.

Your peer is a {agent_domain} specialist. They are strong at {agent_domain} concepts but may make mistakes when the problem involves {peer_domain} reasoning.

Below is the question and your peer's response. Your task has TWO steps.

STEP 1: Anchor to the problem:
Read the QUESTION carefully. Identify ONE key technical term or concept word (1-3 words) that is central to the problem. Use the EXACT wording as it appears in the question text. Do NOT use generic words such as "problem", "question", "answer", "value", "calculation", "formula", "student", "peer", or "specialist" -- these will be rejected. Output the concept in <concept>...</concept> tags BEFORE writing any other text.

STEP 2: Provide feedback:
After the concept tag, write your feedback as a short paragraph of 3 to 5 sentences (no more than about 80 words total). Focus on the {peer_domain} aspects of the reasoning where relevant. Point out what is correct and what is wrong or missing.

Do NOT solve the problem yourself. Cite specific steps or lines in the peer's response. Do NOT directly reveal the correct answer letter.

Question:
{problem}

The {agent_domain} specialist's response:
{response}

Your response (begin with <concept>...</concept>, then a 3-5 sentence paragraph):
\end{lstlisting}
\end{promptbox}

\subsection{Sanitizing Feedback}
\label{append:sanitize_fb}

Before tutor feedback is injected into the tutee's reprompt, we apply a lightweight sanitization and validation pipeline. 
First, we remove explicit answer-revealing patterns, such as boxed answer letters, bolded answer letters, and phrases like ``Final answer: X'' or ``The correct option is X''. 
We do not remove numeric boxed values or ordinary references to answer options inside explanatory text. 
Second, we validate the concept anchor produced by the feedback-anchoring prompt: the extracted \texttt{<concept>...</concept>} field must contain at least one non-generic word that appears in the problem text. 
If this validation fails, the feedback is dropped. 
Finally, for validated feedback, we strip the concept tag before injecting the feedback, so the tutee receives only the feedback content.

\begin{table}[H]
\centering
\small
\resizebox{\linewidth}{!}{%
\renewcommand{\arraystretch}{1.05}
\begin{tabular}{ll}
\toprule
\textbf{Tag} & \textbf{Meaning} \\
\midrule
\texttt{match:<word>} 
& Kept; \texttt{<word>} appears in the problem text. \\
\texttt{no\_concept\_tag} 
& Dropped; \\
& the concept tag is absent or malformed. \\
\texttt{empty} 
& Dropped; \\
& the concept tag is present but empty. \\
\texttt{all\_generic} 
& Dropped; \\
& the concept only contains generic words. \\
\texttt{no\_match} 
& Dropped; \\
& the concept does not appear in the problem. \\
\texttt{fb\_none} 
& Dropped; \\
& feedback generation returned no output. \\
\bottomrule
\end{tabular}
}
\caption{Outcome categories for feedback anchoring process.}
\label{tab:feedback_validation_categories}
\end{table}

\section{Experimental Details}\label{append:experimental_detail}

\subsection{Setup for Multi-Domain Science Q\&A Experiment}
\label{append:science_qa_setup}

We follow the data construction of~\citet{sdpo}.
For each domain in SciKnowEval~\cite{sciknoweval}, we split the data into train and test sets with a 9:1 ratio. 
For OPCoD, we additionally sample a small validation set from the training portion, yielding disjoint train, validation, and test partitions.

To construct the initial students, we start from Qwen3-8B~\cite{yang2025qwen3} and supervised fine-tune a separate model for each domain using chain-of-thought (CoT) data from that domain. 
The resulting students are not intended to be perfect solvers; rather, each student is relatively stronger on its target domain than the other domain-tuned students. 
We use these three single-domain SFT models as the initial students for OPCoD.

Since SciKnowEval provides problem--answer pairs without CoT rationales, we generate rationalized solutions by prompting \texttt{gpt-5-mini} with each training problem and its gold answer. 
SFT is performed with LoRA~\cite{hu2022lora} using rank 8 on all linear layers, learning rate $2\times10^{-4}$ with cosine scheduling and 10\% warmup, batch size 16, three epochs, and bf16 precision.

\subsection{Implementation Details}
\label{sec:impl_details}

We implement GRPO and SDPO using the official codebase of~\citet{sdpo}, utilizing their codebase in full compliance with its Apache-2.0 license for academic research. 
OPCoD is built on top of SDPO: each student is trained with the same self-distillation objective, but the self-teacher is additionally conditioned on tutor feedback. 
Here, the self-anchor is the student's earliest correct rollout for the same problem when available, and is empty otherwise. 
The feedback is generated by the frozen tutor model, then passed through sanitization and feedback anchor validation before being used.

Relative to SDPO, OPCoD adds four implementation components. 
First, a feedback collector runs the tutor model in a separate vLLM instance and generates feedback for the student responses in each training batch. 
Second, concept-anchor validation discards feedback whose extracted \texttt{<concept>} tag does not appear in the problem text, filtering ungrounded feedback. 
Third, sanitization removes direct answer-revealing patterns, such as boxed answer letters, to prevent shortcut imitation. 
Fourth, dynamic cognizance-based gating evaluates both agents on a validation set at the end of each round; if the tutor is classified as incognizant, feedback collection is skipped in the next round and training falls back to self-distillation without feedback.

Each OPCoD round consists of two directional phases. 
In the first phase, one student is updated while the other is loaded as a frozen tutor; in the second phase, their roles are swapped. 
At the end of each phase, we save both FSDP shards and a HuggingFace-merged checkpoint, which is used to load the tutor model for the next phase. 
All experiments are run on a single node with $2\times$ NVIDIA H200 GPUs. 
The actor uses FSDP across the two GPUs, while the tutor vLLM process is colocated on the same GPUs with limited memory utilization.

\subsection{Hyperparameters}\label{append:hyperparams}

We report the main hyperparameters used in our experiments. 
Table~\ref{tab:hyperparams_sdpo} lists the SDPO hyperparameters, which also serve as the base configuration for OPCoD. 
Table~\ref{tab:hyperparams_codistill} reports the additional hyperparameters specific to OPCoD. 
Table~\ref{tab:hyperparams_grpo} lists the GRPO baseline hyperparameters.


\begin{table}[H]
\centering
\small
\renewcommand{\arraystretch}{1.1} 
\setlength{\tabcolsep}{8pt}

\begin{tabularx}{\linewidth}{X l}
\toprule
\textbf{Parameter} & \textbf{Value} \\
\midrule
\multicolumn{2}{l}{\textbf{General}} \\
\midrule
Model & Qwen/Qwen3-8B \\
Thinking & False \\

\midrule
\multicolumn{2}{l}{\textbf{Data}} \\
\midrule
Max prompt length & 2048 \\
Max response length & 4096 \\
Max model length & 18944 \\

\midrule
\multicolumn{2}{l}{\textbf{Batching / Rollout}} \\
\midrule
Question batch size & 32 \\
Mini batch size & 32 \\
Number of rollouts & 8 \\
Inference engine & vLLM \\
Temperature / Top-$p$ & 1.0 / 1.0 \\

\midrule
\multicolumn{2}{l}{\textbf{Evaluation}} \\
\midrule
Number of rollouts & 16 \\
Temperature / Top-$p$ & 0.6 / 0.95 \\

\midrule
\multicolumn{2}{l}{\textbf{SDPO loss}} \\
\midrule
Full-logit distillation & True \\
Top-$K$ distillation & 100 \\
Tail bucket & True \\
Distillation divergence & JSD ($\alpha=0.5$) \\
Teacher regularization & EMA \\
Teacher EMA update rate & 0.05 \\
Rollout correction & Token-level IS, threshold 2.0 \\

\midrule
\multicolumn{2}{l}{\textbf{Training}} \\
\midrule
Optimizer & AdamW \\
Learning rate & $1\times10^{-5}$, constant \\
Warmup steps & 5 \\
Weight decay & 0.01 \\
Gradient clip norm & 1.0 \\
Total steps per agent & 100 \\
\bottomrule
\end{tabularx}
\caption{Hyperparameters for the SDPO.}
\label{tab:hyperparams_sdpo}
\end{table}

\begin{table}[h]
\centering
\small
\renewcommand{\arraystretch}{1.1} 
\setlength{\tabcolsep}{8pt}

\begin{tabularx}{\linewidth}{X l}
\toprule
\textbf{Parameter} & \textbf{Value} \\
\midrule
\multicolumn{2}{l}{\textbf{Training}} \\
\midrule
\# rounds & Phys--Chem (2)\\
  & Chem--Mat (2) \\
  &  Mat--Phys (5) \\
Total steps per agent & 100 \\

\midrule
\multicolumn{2}{l}{\textbf{Data}} \\
\midrule
Peer inference backend & vLLM \\
Max tokens for feedback & 1024 \\
Max model length & 8192 \\
Max reprompt length & 10240 \\

\midrule
\multicolumn{2}{l}{\textbf{Feedback processing and gating}} \\
\midrule
Threshold $\tau$ & 0.2 \\
Validation set & 60 problems \\
  & balanced per domain \\
Samples per validation problem & 16 \\
\bottomrule
\end{tabularx}
\caption{Additional hyperparameters for OPCoD. The underlying SDPO hyperparameters are the same as in Table~\ref{tab:hyperparams_sdpo}.}
\label{tab:hyperparams_codistill}
\end{table}

\section{Break-Rate Analysis}
\label{append:break-risk-appendix}
We diagnose whether tutor feedback can change the correctness of the tutee's response. 
For each domain pair, we take the two initial students and evaluate them on a held-out subset of the pair's training distribution. 
For each problem, a student first generates $n=8$ rollouts. 
The other student then generates feedback, which is injected back into the reprompt together with the original problem and previous correct rollout if exists.
The same student then re-answers the problem, which takes a role of self-teacher.
For every rollout, we record whether the pre-feedback answer is correct or wrong, and whether the post-feedback answer is correct or wrong. 
The \emph{break-rate} is the fraction of originally correct rollouts that become wrong after adding feedback:
\[
\mathrm{BreakRate}
=
\frac{\#(C \rightarrow W)}{\#C}.
\]
We group problems by difficulty using the number of wrong pre-feedback rollouts among the eight sampled rollouts: very easy (0--1), easy (0--4), hard (5--7), and hardest (7).
Across the three domain pairs, the aggregate break-rate is 2.77\% for cognizant tutors and 6.53\% for incognizant tutors; restricting to problems in the tutor's stronger domain gives 3.56\% and 5.12\% respectively.

\section{Additional Case Studies}\label{append:additional_examples}

We present additional examples demonstrating how tutor feedback effectively guides the tutee in resolving diverse conceptual errors.

\begin{figure*}[t]
    \centering
    \includegraphics[width=\linewidth]{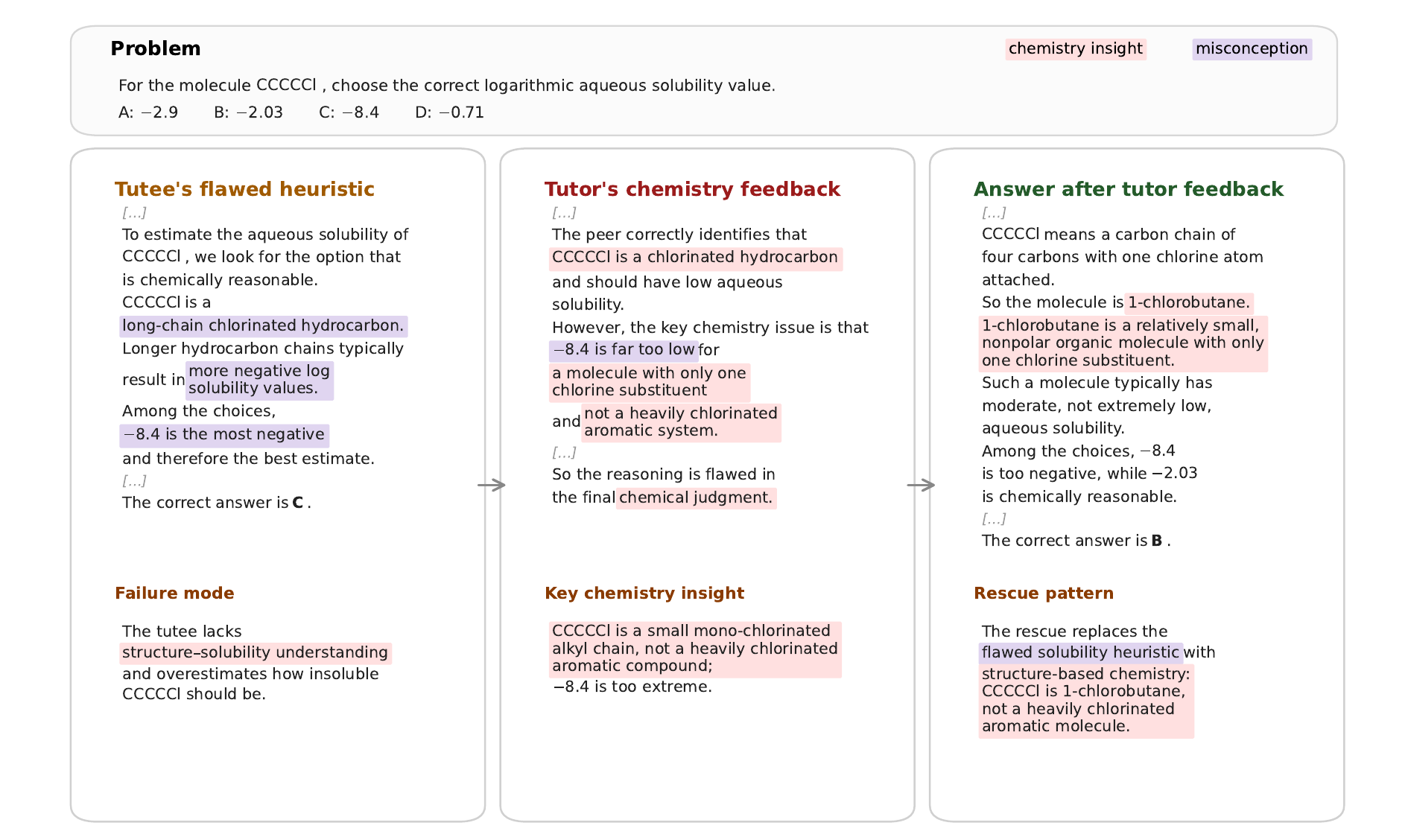}
    \caption{Chemistry-domain example where tutor feedback corrects a chemistry misconception in the tutee's original reasoning.}
\label{fig:case_study_chem_1039}
\end{figure*}

Figure~\ref{fig:case_study_chem_1039} shows a chemistry-domain example where the tutee's failure comes from a chemistry misconception rather than from being stuck in physics-style reasoning. 
The chemistry-stronger tutor corrects this misconception, allowing the tutee to revise its answer.

\begin{figure*}[t]
    \centering
    \includegraphics[width=\linewidth]{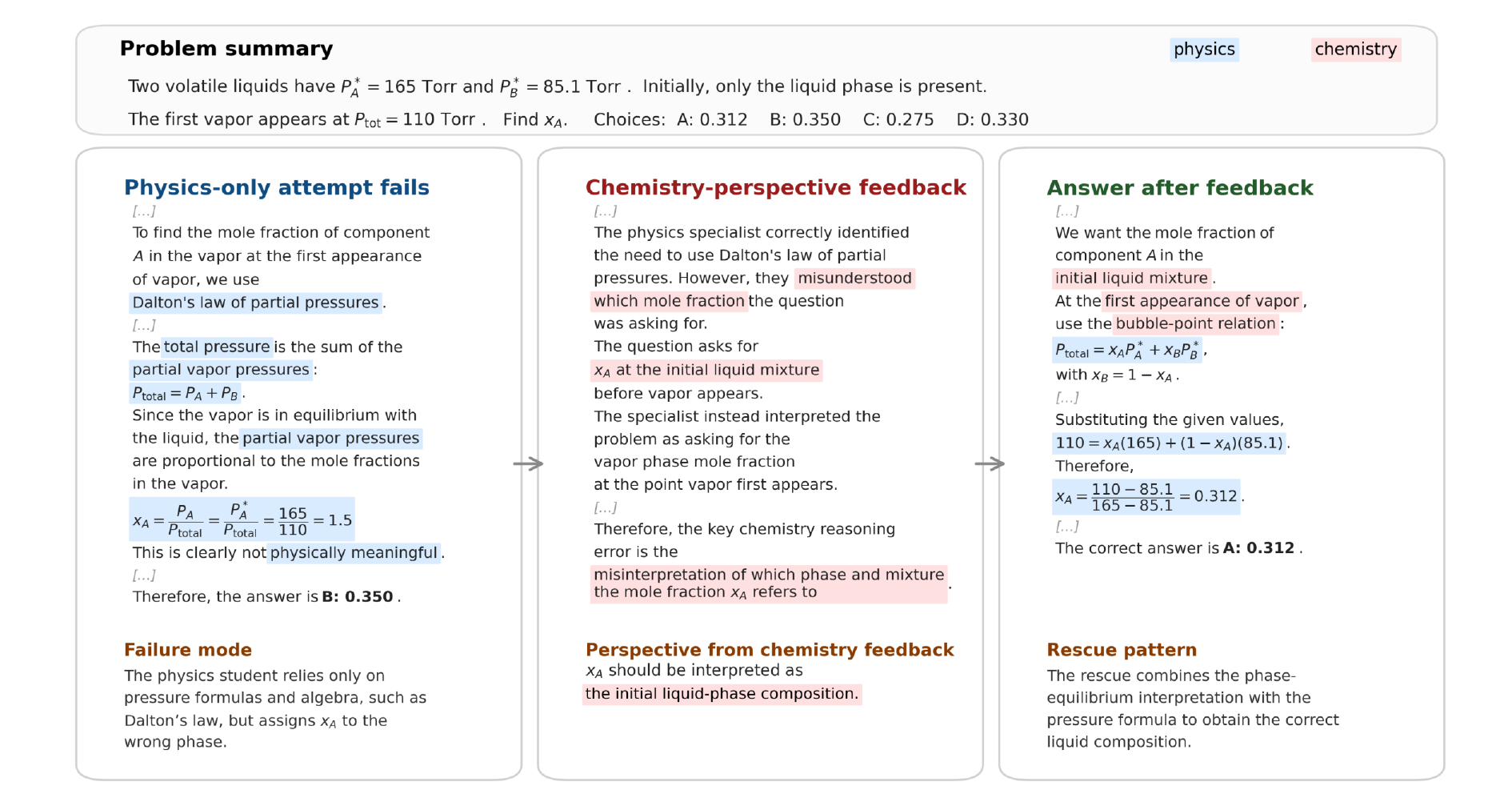}
    \caption{Physics-domain example where chemistry-perspective feedback helps the tutee identify the missing phase-equilibrium interpretation and revise its answer.}
\label{fig:case_study_phys_190}
\end{figure*}

Figure~\ref{fig:case_study_phys_190} shows a physics-domain example where the physics-stronger tutee applies relevant formulas but misses the phase-equilibrium interpretation of the quantity being asked. 
The chemistry-stronger tutor points out this missing interpretation, enabling the tutee to combine it with its physics calculation and correct the answer.

\begin{table}[t]
\centering
\small
\renewcommand{\arraystretch}{1.1} 
\setlength{\tabcolsep}{8pt}

\begin{tabularx}{\linewidth}{X l}
\toprule
\textbf{Parameter} & \textbf{Value} \\
\midrule
\multicolumn{2}{l}{\textbf{General}} \\
\midrule
Model & Qwen/Qwen3-8B \\
Thinking & False \\

\midrule
\multicolumn{2}{l}{\textbf{Data}} \\
\midrule
Max prompt length & 2048 \\
Max response length & 8192 \\
Max model length & 10240 \\

\midrule
\multicolumn{2}{l}{\textbf{Batching / Rollout}} \\
\midrule
Question batch size & 32 \\
Mini batch size & 32 \\
Number of rollouts & 8 \\
Inference engine & vLLM \\
Temperature / Top-$p$ & 1.0 / 1.0 \\

\midrule
\multicolumn{2}{l}{\textbf{Evaluation}} \\
\midrule
Number of rollouts & 16 \\
Temperature / Top-$p$ & 0.6 / 0.95 \\

\midrule
\multicolumn{2}{l}{\textbf{Algorithm}} \\
\midrule
Normalize advantage by std & False \\
Critic & Disabled \\
KL loss & Disabled \\
Rollout correction & Token-level IS, threshold 2.0 \\

\midrule
\multicolumn{2}{l}{\textbf{Training}} \\
\midrule
Total steps per agent & 100 \\
Optimizer & AdamW \\
Learning rate & $1\times10^{-5}$, constant \\
Warmup steps & 5 \\
Weight decay & 0.01 \\
Gradient clip norm & 1.0 \\
\bottomrule
\end{tabularx}
\caption{Hyperparameters for the GRPO.}
\label{tab:hyperparams_grpo}
\end{table}

\section{Walltime Analysis}\label{append:walltime}

\begin{figure}[H]
    \centering
    \includegraphics[width=\linewidth]{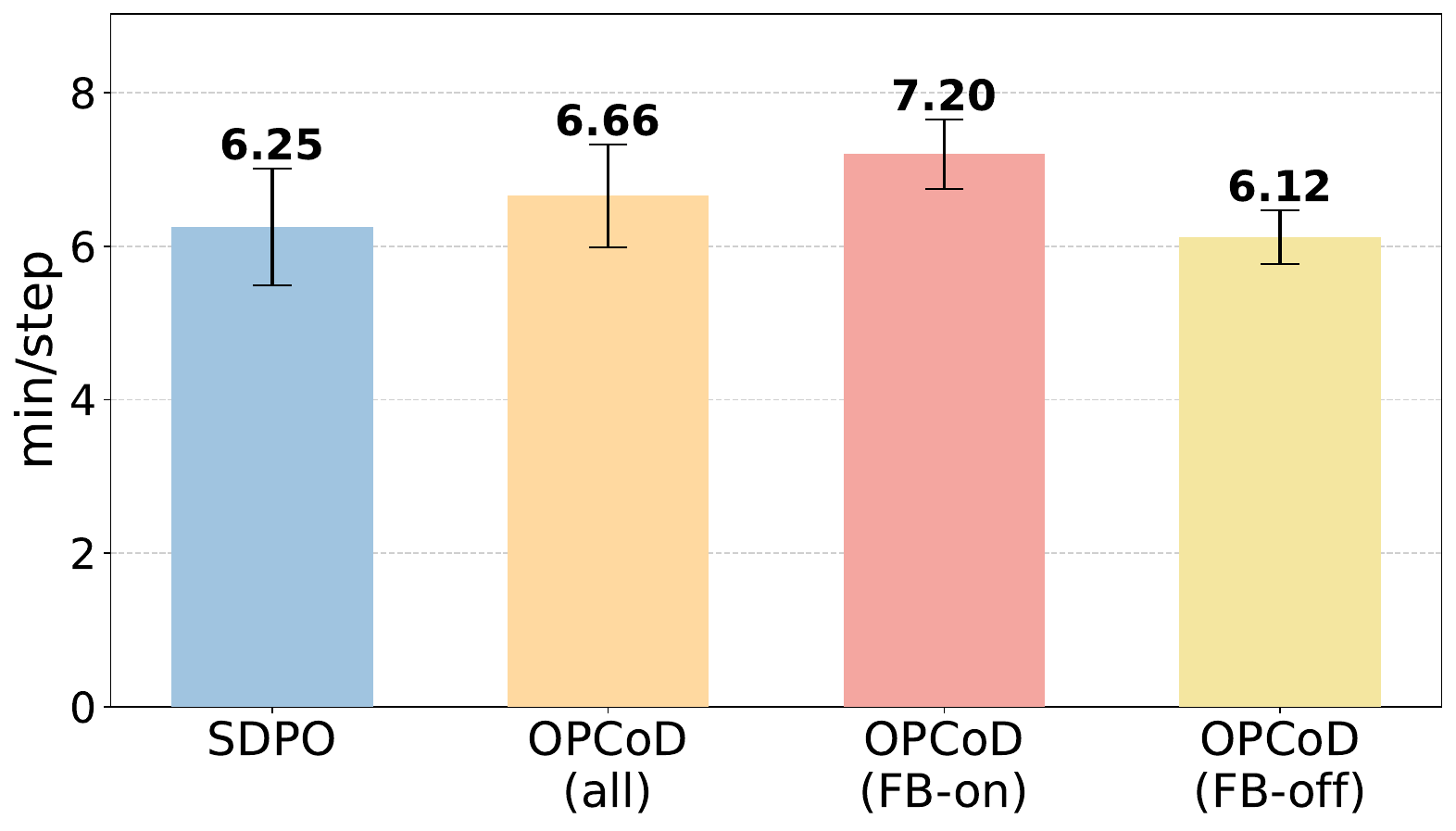}
    \caption{Average walltime per training step on the physics--chemistry experiment. FB-on and FB-off denote OPCoD phases with and without feedback generation, respectively.}
\label{fig:walltime_analysis}
    \label{fig:walltime_m2_sdpo_pesd_fb_split}
\end{figure}

OPCoD adds tutor-feedback generation on top of the SDPO training pipeline, which can introduce additional walltime cost; we analyze this cost on the physics--chemistry experiment. 
Figure~\ref{fig:walltime_analysis} reports the average walltime per training step. 
We report the overall OPCoD average, and also separate OPCoD steps where feedback generation is enabled (FB-on) from those where feedback is skipped by the gating mechanism (FB-off).

The overall per-step walltime increases only modestly from 6.25 min/step for SDPO to 6.66 min/step for OPCoD. 
As expected, FB-on steps are slower due to feedback generation, while FB-off steps have a cost comparable to SDPO. 
Thus, the feedback mechanism introduces a limited overhead in the overall training pipeline.

\clearpage

\end{document}